\documentclass[letterpaper]{article}
\usepackage{aaai16}
\usepackage{times}
\usepackage{helvet}
\usepackage{courier}
\usepackage{graphicx}
\usepackage{comment}
\usepackage{url}      
\usepackage{tabularx}
\usepackage{amssymb}
\usepackage{booktabs}
\usepackage{balance}  
\usepackage{wasysym}

\begin{document}
%
\title{Shirtless and Dangerous: Quantifying Linguistic Signals of Gender Bias\\ in an Online Fiction Writing Community}
\author{
 Ethan Fast, Tina Vachovsky, Michael S. Bernstein \\
 Stanford University\\
 \{ethan.fast, tvachov, msb\}@cs.stanford.edu
\\
}
\maketitle
\begin{abstract}
Imagine a princess asleep in a castle, waiting for her prince to slay the dragon and rescue her. Tales like the famous \textit{Sleeping Beauty} clearly divide up gender roles. But what about more modern stories, borne of a generation increasingly aware of social constructs like sexism and racism? Do these stories tend to reinforce gender stereotypes, or counter them? In this paper, we present a technique that combines natural language processing with a crowdsourced lexicon of stereotypes to capture gender biases in fiction. We apply this technique across 1.8 billion words of fiction from the Wattpad online writing community, investigating gender representation in stories, how male and female characters behave and are described, and how authors' use of gender stereotypes is associated with the community's ratings. We find that male over-representation and traditional gender stereotypes (e.g., \textit{dominant} men and \textit{submissive} women) are common throughout nearly every genre in our corpus. However, only some of these stereotypes, like \textit{sexual} or \textit{violent} men, are associated with highly rated stories. Finally, despite women often being the target of negative stereotypes, female authors are equally likely to write such stereotypes as men. 
\end{abstract}

\section{Introduction}

\begin{quote}
\footnotesize
''Ooh! Ooh! Look at this top! I have to have it.'' Nina said holding up a gold, glittery top with a boat neck but low cut back. It was gorgeous. ''And those shoes!''

-- \textit{Anonymous} (a Wattpad story)

"I highly doubt that," Fanny growled, clenching a hammer in her hands that she pulled from her work belt.

-- \textit{Anonymous} (a Wattpad story)

\normalsize
\end{quote}

These two excerpts from stories on an online writing community lay out competing concepts of gender roles in fiction. The first plays on common female stereotypes, while the second rejects them. Fictional worlds offer writers open-ended laboratories where they might explore new forms of social norms. Do these worlds tend to reinforce stereotypes, or push for more balanced gender roles? This paper is about quantitatively untangling signals of gender bias in an online community of amateur fiction writers.
%
%
%

Gender bias and its stereotypes retain significant force in modern culture. The idea of domestic and dependent women, for example, lives on in Hollywood and through subtle aspects of discrimination in science, C-level industry, and the broader workplace \cite{science,industry}. Similarly, the stereotype of violent and sexual men is abundant in television, video games, and popular culture \cite{roleplayinggames,music}. At the more basic level of gender representation, men are over-represented in news coverage and television shows, among other media \cite{news,primetime}.

Works of online fiction allow us to detect these gender stereotypes and others at scale, by examining how web users write male and female characters. The fact that such authors are \textit{amateurs} helps to minimize the market incentives that drive commercial novels. What norms emerge outside these constraints? To find out, we analyze more than 1.8 billion words of fiction on Wattpad, an online community where millions of people share stories that they have written. 

Four research questions guide our analysis, collectively  aiming to investigate how male and female characters behave, how they are portrayed, and what reflection those characterizations have on the community:

\textbf{RQ1-Action: } What role does gender play in how characters act? Are male and female characters associated with different kinds of verbs?
%
%
%

\textbf{RQ2-Description:} What role does gender play in how characters are described? Are male and female characters associated with different kinds of adjectives? 

\textbf{RQ3-Ratings:} How do patterns in character action and description impact a story's rating? Are stories that deploy 
stereotypes more highly rated?

\textbf{RQ4-Authors:} How does \textit{author} gender impact the the stereotypes we discover? Do male authors write women differently than female authors, and vice versa?
%
%
%

We measure gender bias by investigating stories at the level of sentences. Sentences provide a rich set of low level signals (pronoun-associated verbs and adjectives) that relate gender to character action and description. To more easily interpret these signals, we use paid crowdsourcing to group common verbs and adjectives into sixteen high level categories like ``sexual'' and ``submissive'' that capture  dimensions of gender stereotypes. Over billions of words, clear patterns emerge that separate male and female characters within these categories. For example, we find that male characters are more physically \textit{active} and \textit{violent}, while female characters are described by terms that have \textit{submissive} or \textit{childish} connotations.   

The size and genre variety of our corpus allow us to probe further into more nuanced questions about these stereotypes. Is there any difference in the way male and female authors write gender stereotypes? We might imagine, for example, that female authors would have a different conception of gender roles than male authors, and so write their characters differently. 
And what about the reception of these stereotypes? Are they associated with higher or lower rated stories?
Ultimately, we find that while both genders perpetuate the same set of stereotypes on Wattpad, the relationship of these stereotypes to story rating is mixed.

\section{Related Work}
\textbf{Gender Stereotypes in Modern Storytelling}: Studies of gender roles in commercial media offer a useful starting point for investigating stereotypes in amateur fiction. Work analyzing movies, children's books, and music lyrics has found that women are likely to be portrayed as younger, more emotionally motivated, less rational, and valued for beauty over intellect \cite{disneyfilms,children}, whereas men are stronger, more violent, less in control of their sexuality, and expected to express emotion physically \cite{dutchtv,music}. Women also tend to be portrayed as domestic, with activities centered around the home or family, while men are shown as physically active and economically successful, with a much wider range of employment and often higher ranking and higher paying jobs \cite{primetime,wrestling}. Where past studies have focused on gender bias in relatively small datasets of professionally-crafted commercial media, our work addresses the biases of amateurs across orders of magnitude more user-generated data.

\textbf{Gender Stereotypes on the Web}: A second body of work has found that similar gender stereotypes translate to the web. For example, males are over-represented in the reporting of web-based news articles \cite{newsimages}, an effect which also holds for other forms of social media, like twitter conversations \cite{bechdel}. Gender roles can also shape the substance of web content itself. For example, biographical articles about women on Wikipedia disproportionately discuss romantic relationships or family-related issues \cite{wikipedia}. Sometimes gender can drive user interaction: for example, on Pinterest, women earn more repins, but fewer followers \cite{pinterest}; on IMDB, reviews written by women are perceived as less useful \cite{imdb}; or on social media sites and dating profiles, women are more often judged on physical characteristics than men \cite{facebookpictures,descriptionbias,datingprofiles}. Similarly, in web searches for images of archetypal occupations (e.g., nurse, investment banker), the minority gender is often portrayed and perceived as relatively more unprofessional \cite{imagesearch}. Our work extends these analyses to the fictional worlds created by a community of amateur writers.


\textbf{Gender and User Behavior}: The gender of a user or artist can influence how they behave or what kinds of stereotypes occur in creative content. For example, on Pinterest, women more often reciprocate social links, while men specialize in curating specific categories of content \cite{pinterest2}. Similarly, in roleplaying games, both genders are likely to choose warriors as novices, but expert female players are more likely to default to the stereotypical healer; men, on the other hand, mainly play healers only if they design the avatar as a woman \cite{roleplayinggames}. In television, the gender of writers can affect plot and narrative: mixed gendered groups of writers are more likely to deviate from stereotypes, writing both men and women in stereotypical feminine roles \cite{primetime}. In our work, we examine how the gender of users on Wattpad affects the kinds of characters that appear in their stories.

\textbf{Analysis of Text Corpora}: Our approach to text analysis is informed by a large body of work in the computational social sciences. Researchers often use lexicons like LIWC to analyze texts across a broad range of signals \cite{liwc}. Other researches have crowdsourced new lexical categories when they do not yet exist \cite{wikipedia,empath}, an approach we adopt to build categories that capture common gender stereotypes. To extract relevant signals from text, researchers have leveraged the coefficients of regression models \cite{enron} and the strength and significance of correlations \cite{wisdom}, as well as performed significance tests between the means of lexical categories \cite{contagion}. In our work, we extend these analyses to target sentence-level relationships that capture behavioral and descriptive attributes of male and female characters.

\begin{table*}[!ht]\scriptsize
  \renewcommand{\arraystretch}{1.2}
  \begin{tabular}{p{2.3cm}p{1.1cm}p{1.1cm}p{1.4cm}p{1.4cm}p{1.5cm}p{1.45cm}p{1.4cm}p{1.6cm}}
    \textbf{Genre} & \textbf{Stories} & \textbf{Rating} & \textbf{M. Authors} & \textbf{F. Authors} & \textbf{M/F Authors} & \textbf{M. Pronouns} & \textbf{F. Pronouns} & \textbf{M/F Pronouns} \\
    \hline
    Historical Fiction & 1562 & 2.49 & 748 & 797 & 0.94 & 0.0098 & 0.0091 & 1.08 \\
    Vampire & 6936 & 2.92 & 3423 & 3466 & 0.99 & 0.0111 & 0.0089 & 1.25 \\
    Teen Fiction & 109873 & 2.84 & 56771 & 52472 & 1.08 & 0.0094 & 0.0084 & 1.12 \\
     Other & 70423 & 3.03 & 38864 & 31109 & 1.25 & 0.0081 & 0.0071 & 1.14 \\
    Fantasy & 21351 & 2.54 & 10951 & 10282 & 1.07 & 0.0092 & 0.0095 & 0.97 \\
    Romance & 70354 & 2.89 & 36471 & 33455 & 1.09 & 0.0110 & 0.0085 & 1.29 \\ 
    Science Fiction & 6193 & 2.63 & 3197 & 2964 & 1.08 & 0.0092 & 0.0070 & 1.31 \\
    Fanfiction & 228707 & 2.90 & 125525 & 101789 & 1.23 & 0.0110 & 0.0076 & 1.45 \\
    Humor & 11472 & 2.71 & 5982 & 5436 & 1.10 & 0.0065 & 0.0057 & 1.14 \\
    Mystery / Thriller & 10164 & 2.77 & 5128 & 4966 & 1.03 & 0.0096 & 0.0093 & 1.03 \\
    Horror & 7287 & 2.95 & 3815 & 3431 & 1.11 & 0.0091 & 0.0093 & 0.98 \\
    Adventure & 11423 & 2.43 & 6088 & 5268 & 1.16 & 0.0087 & 0.0082 & 1.06 \\
    Paranormal & 5603 & 2.80 & 2872 & 2696 & 1.07 & 0.0091 & 0.0092 & 0.99 \\
    Spiritual & 1477 & 2.38 & 788 & 676 & 1.17 & 0.0073 & 0.0071 & 1.03 \\
    Action & 5901 & 2.82 & 3175 & 2691 & 1.18 & 0.010 & 0.0072 & 1.39 \\
    Non-Fiction & 8258 & 2.46 & 4390 & 3822 & 1.15 & 0.0053 & 0.0054 & 0.98 \\
    Short Story & 11991 & 2.50 & 6353 & 5567 & 1.14 & 0.0087 & 0.0086 & 1.01 \\
    General Fiction & 5290 & 2.87 & 2830 & 2431 & 1.16 & 0.0089 & 0.0090 & 0.99 \\
    Werewolf & 12528 & 2.96 & 6453 & 6000 & 1.08 & 0.011 & 0.0082 & 1.22 \\
    ChickLit & 2031 & 2.93 & 1095 & 923 & 1.19 & 0.0091 & 0.0087 & 1.05 \\
    \hline
    \textbf{All} & 655295 & 2.70 & 349795 & 301530 & 1.16 & 0.0096 & 0.0077 & 1.24 \\
    \hline
  \end{tabular}
  \caption{An overview of the genres in our fiction dataset, along with high-level gender statistics. Male characters are generally over-represented across genres. The \textit{Stories} header refers to the number of stories per genre. \textit{Rating} refers to the average rating of stories. \textit{M. Authors} and \textit{F. Authors} refer to the number of stories written by men and women. \textit{M/F Authors} refers to the ratio of male to female authors. \textit{M. Pronouns} and \textit{F. Pronouns} refer to the average number of male and female pronouns per story, normalized by story length. \textit{M/F Pronouns} refers to the ratio of male to female pronouns.}
  \label{tbl:rep-stats}
\end{table*}

\section{Data and Methods}
We take a quantitative approach to answer our research questions. Here we describe our data and statistical methods.

\subsection{Data: 1.8 billion words of amateur fiction}
We conducted our analyses over amateur fiction from the Wattpad writing community.\footnote{\url{http://wattpad.com}} In aggregate, our dataset contains more than 1.8 billion words selected from a random sample of 600,000 stories, written by more than 500,000 writers across twenty genres. Wattpad provided us with all data, as well as gender annotations for 475,000 authors, 46\% of which are by women and 54\% of which are by men. Table \ref{tbl:rep-stats} presents a summary of the stories and genres in our dataset, as well as the gender representation in each genre, captured by counts of male and female pronouns. 
%
%
%
%
%
%

How balanced are Wattpad stories between male and female characters?
Before addressing our research questions, it is important to note (and control for) the fact that men are commonly over-represented in many forms of media \cite{news,bechdel}. This representational bias also exists on Wattpad (Table \ref{tbl:rep-stats}). Consistent with prior work, male pronouns are more common across most genres: Wattpad stories use 1.24 times more male pronouns in their writing than female pronouns. 
Further, both genders write more about men. Male authors use 1.23 times more male than female pronouns, and female authors use 1.25 times more male pronouns (a small but significant difference, $p < 0.001$). Our methodology adjusts for this male over-representation by normalizing our statistics between genders. 

\subsection{Character Statistics}
Social scientists often treat documents as ``bags of words,'' for example analyzing tweets, books, or status updates with the lexical categories in tools like LIWC \cite{contagion,wisdom}. Here we present a more targeted kind of analysis that captures the actions and descriptions of characters by taking advantage of the dependency structure of sentences.

\textbf{Character behavior.} Stereotypes are defined in part by how people behave. Do our stories read like Indiana Jones movies, with men initiating violence and activity, and women screaming and needing rescue? To address these kinds of questions, we mined gendered \textit{character actions} from the dataset.

We extracted actions through subject-verb relationships where the subject is ``he'' or ``she''. For example, in the sentence ``she ruined him utterly'' we would extract the subject-verb relationship ``she ruins''. We lemmatized words in these extractions (e.g., ``ruined'' to ``ruins'') to collapse verbs onto a canonical form. 

This process left us with 42 million character actions, of which 25 million are male, and 17 million are female.







\textbf{Character description. } Stereotypes also exist through how people are described. Are men expected to be tall and broad-shouldered, while women are small and graceful, as cultural stereotypes would suggest? To find out, we mined gendered \textit{adjective descriptors} from the dataset.

We extracted two kinds of adjective relationships. First, we targeted adjectives attached to male or female subjects by the verb ``to be,'' for example ``he was clinically \textit{deranged}.''  Second, we targeted the modifiers of words that represent male or female characters like ``brother'' or ``mother,''  for example ``the \textit{upset} mother.'' Again, we lemmatized all words. 

This process left us with 4.3 million adjectives, of which 2 million are applied to female characters, and 2.3 million are applied to male characters.

\subsection{Crowdsourcing a Stereotype Lexicon}
Character statistics present a low level view of the differences between genders. To understand stereotypes at a higher level -- for example, whether male characters more often act \textit{violently} or female characters are more often described as \textit{weak} -- we need lexical categories that map individual words onto broader categories. Such categories would tell us that the adjective ``fragile'' is associated with  \textit{weak}, or the verb ``kill'' is associated with \textit{violence}. 

To choose these categories of stereotype, we draw from the results of prior work across television, movies, and social media \cite{disneyfilms,primetime}. The literature identifies the following set of categories predominantly associated with men: \textit{violent}, \textit{dominant}, \textit{strong}, \textit{arrogant}, \textit{sexual}, \textit{angry}, and (physically) \textit{active}. For women, the predominant categories are: \textit{domestic}, \textit{hysterical}, \textit{childish}, \textit{afraid}, \textit{dependent}, \textit{emotional}, \textit{beautiful}, and \textit{submissive}. We utilize these categories in our analysis.

We built lexical categories that capture this list of stereotypes by mapping the 2000 most commonly occurring verbs and adjectives in our dataset onto the set of categories through a series of crowdsourcing tasks. Labeling the most common words in a corpus is an efficient way to build a useful lexicon while limiting cost \cite{emolex}. For each  task, we asked workers on Amazon Mechanical Turk (AMT):

\begin{quote}
For each word, tell us how strongly it relates to the topic. For example, given the topic VIOLENT, the words ``kill'' and``hurt'' would be \textit{strongly related}, the word ``scratch'' would be \textit{related}, the word ``resist'' would be \textit{weakly related}, and the words ``laugh'' and ``patient'' would be \textit{unrelated}.
\end{quote}

We created tasks for each category of stereotype. Each task gathers annotations for 20 words, for which we pay workers \$0.14 in line with guidelines for ethical research \cite{dynamo}. We collect labels from three independent workers, enough to generate accurate results given high quality workers, such as the Masters workers we recruit on AMT \cite{get-another-label}.
%
%
%
To reduce the impact of false positives on our analysis, we added a word to a category of stereotype only if the majority of workers have decided it is related or strongly related to the category in question.
Similarly defined tasks have preformed well when constructing lexical categories \cite{emolex}.

We generated this lexicon at a total cost of \$630. Crowd workers agreed among themselves (voting unanimously for labels of related or strongly related) at a rate of 85\%.

\subsection{Statistical Methods}

Given a lexicon that maps low level character statistics onto high level stereotypes, we now present the statistical techniques that translate these data into answers to our research questions. We used two primary methods to investigate our four research questions:

\textbf{RQ1-Action and RQ2-Description}. To discover the presence of gender stereotypes in how characters act and are described, we tested for a difference in mean frequency counts between genders for 
our sixteen stereotype categories,
taking these means over stories and normalizing them on character gender occurrence rates. We used Welch's t-test, a two-sided test that does not assume equal variance among populations, and we present these results as odds ratios between means.
%
%
%
%
%
%
Because we investigated many hypotheses at once, we applied a Bonferroni correction ($\alpha = 0.05 / 16 = 0.0031$) to the results of each analysis. All the results we present reach this level of significance. 

To better explain our results, and discover specific behaviors (verbs) and descriptions (adjectives) that drive our high level effects, we also tested for differences in the gendered frequency counts of 2000 individual words within our stereotype categories (e.g., how often the adjective ``cocky'' in the category \textit{dominant} modifies male vs. female pronouns). For these word-level statistics, we again compared normalized mean frequency counts using Welch's t-test and applied a Bonferroni correction ($\alpha = 0.05 / 2000 = 2.5e^{-5}$).

We repeated these statistical tests within individual genres to add nuance to our results and ensure that our findings were not overly influenced by one particularly large genre, like Teen Fiction.

\textbf{RQ3-Ratings and RQ4-Authors}. To discover how stereotypes differ between male and female authors or across high and low rated stories, we trained logistic regression models that use gendered counts from our sixteen stereotype categories to predict either author gender or story rating. This type of model has provided insight into other kinds of linguistic signals \cite{enron}. For example, if men and women write different kinds of stereotypes (e.g., if women write more \textit{hysterical} male characters, or less of them, or follow any other consistent pattern), we would expect a classifier trained on the frequency counts of stereotypes captured by our categories to have significant predictive value on author gender. Similarly, a classifier trained on these features to predict story rating will reveal what stereotypes (e.g., \textit{violent} men or \textit{dominant} women) are associated with higher and lower rated stories.
%
%
%

Concretely, we trained these logistic regression models using scikit-learn with L2 regularization (C=1). For both models, the input features were frequency counts of words captured by our stereotype categories for male and female characters, normalized on the baseline character occurrence rates for each gender.  This made $16*2=32$ features in total. To predict author gender, we trained on two classes (male and female) over a dataset of 429,582 stories. For story rating, we split the dataset into two equal parts \textit{above} and \textit{below} the median rating, and trained the model to predict these classes. When reporting on classifier accuracy, we ran both models under 10-fold cross-validation. We applied a Bonferroni correction ($\alpha = 0.05 / 32 = 0.0016$) when testing the significance of features in these models.

Finally, we computed gendered odds ratios on the 1000 most common verbs and adjectives for both male and female authors. These ratios tell us, for example, how much more likely a male character written by a man will take an action like \textit{punch}, in comparison to a male character written by a women. We plotted a log-transform of these ratios for visual clarity, and computed a Pearson correlation to quantify the similarity between authors of different genders.

\begin{figure*}[!t]
\centering
\includegraphics[width=2.0\columnwidth]{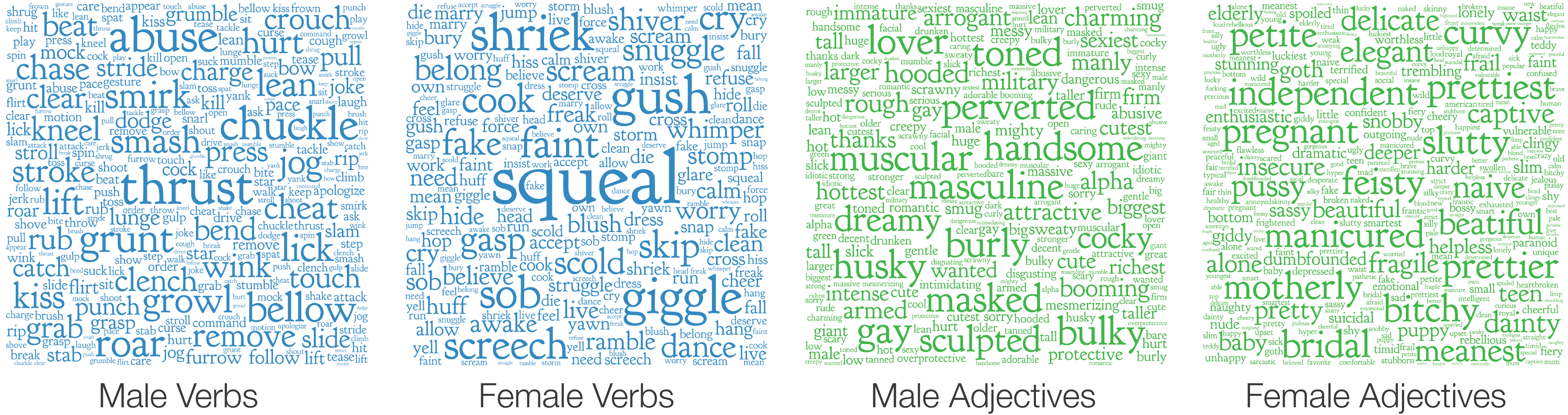}
\caption{Here we present verbs and adjectives that are significantly associated ($p < 2.5e^{-5}$) with male and female characters. The size of each word in the cloud is proportional to its odds ratio for the associated gender.}
\label{fig:wordcloud}
\end{figure*}

\section{Results}

In this section we present the results of our analyses. We describe our raw statistical results here, then elaborate on them with examples in the Discussion section.

\begin{table}[tb]\scriptsize
\renewcommand{\arraystretch}{1.2}
  \begin{tabular}{p{5em}@{\hspace{1.25em}}p{2em}@{\hspace{1.5em}}p{23.25em}}
  \textbf{Male} & \textbf{Odds} & \textbf{Sample verbs and adjectives (male odds)} \\
  \hline
strong & 2.02 & intense (3.1), smash (2.6), intimidating (2.1)\\
arrogant & 1.30 & cocky (7.1), smirk (2.8), smug (2.6), rude (1.4) \\
sexual & 1.22 & sexiest (3.1), kiss (2.4), hot (2.1), flirt (1.5) \\ 
active & 1.17 & jog (2.5), lift (2.4), dodge (1.7), spin (1.4) \\
dominant & 1.15 & rich (2.8), protective (2.7), royal (2.0), command (1.4)\\
violent & 1.10 & abuse (4.4), hurt (2.3), beat (2.0), kill (1.5) \\
beautiful & 1.06 & dreamy (8.14), attractive (4.09), cute (3.3), hot (2.14)\\
angry & 1.05 & bellow (3.1), growl (2.7), curse (1.4), snarl (1.3) \\ 
  \\
  \textbf{Female} & \textbf{Odds} & \textbf{Sample verbs and adjectives (female odds)} \\
  \hline
weak & 1.73 & fragile (6.3), faint (3.2), sick (1.8), tired (1.4) \\
submissive & 1.66 & helpless (3.5), shy (2.9), timid (2.8), whimper (1.7) \\
childish & 1.54 & squeal (11.1), naive (7.8), giggle (4.9), silly (1.7) \\
afraid & 1.46 & shriek (4.8), frightened (2.3), shiver (1.8)\\
dependent & 1.43 & clingy (3.2), vulnerable (2.5), desperate (1.8)\\
hysterical & 1.25 & bitchy (11.4), dramatic (3.2), suicidal (3.1) \\
domestic & 1.16 & cook (2.3), wash (1.8), marry (1.7), clean (1.5)\\
emotional & 1.04 & meanest (7.9), gush (5.1), sob (3.7), fiery (2.8) \\
  \end{tabular}
\caption{The sixteen crowdsourced stereotype categories we used to analyze the corpus. All category odds ratios are statistically significant after correction ($p < 0.0031$). Men are portrayed as more strong, arrogant, and sexual, and women as more weak, submissive, and childish.}
  \label{tbl:code-stats}
\end{table}

\begin{table}[tb]\scriptsize
\renewcommand{\arraystretch}{1.2}
  \begin{tabular}{p{5em}@{\hspace{2em}}p{4em}@{\hspace{2em}}p{19em}}
  \textbf{Stereotype} & \textbf{Direction} & \textbf{Genres that disagree with direction} \\
  \hline
arrogant & male & Non-Fiction \\
sexual & male & General Fiction, Science Fiction, Historical Fiction, Adventure \\ 
angry & male & Fantasy, Chicklit, Teen Fiction \\ 
domestic & female & Spiritual \\
emotional & female & Science Fiction, Historical Fiction, Adventure, Fanfiction, Action, Spiritual 
\end{tabular}
\caption{Stereotypes are mostly consistent across genres. Here we present the few stereotype categories that show some disagreement within genres. \textit{Direction} refers to the gender association of each stereotype on the whole dataset. This effect runs in the opposite direction for genres that disagree. E.g., in Chicklit women are more associated with anger than men.}
  \label{tbl:by-genre}
\end{table}

\subsection{RQ1 and RQ2: Behavior and Representation}
All of our stereotype categories are statistically associated with either male or female characters, and these effects are significant after correction (Table \ref{tbl:code-stats}). 
We find men are more often
\textit{strong} (2.02 odds),
\textit{arrogant} (1.30 odds),
\textit{sexual} (1.22 odds),
\textit{active} (1.17 odds),
\textit{dominant} (1.15 odds),
\textit{violent} (1.10 odds),
\textit{beautiful} (1.06 odds),
and \textit{angry} (1.05 odds).
Women are more often
\textit{weak} (1.73 odds),
\textit{submissive} (1.66 odds),
\textit{childish} (1.54 odds),
\textit{afraid} (1.46 odds),
\textit{dependent} (1.43 odds),
\textit{hysterical} (1.25 odds),
\textit{domestic} (1.16 odds),
and
\textit{emotional} (1.03 odds). 

When we analyze the same categories within individual genres, we mostly find consistent agreement with the overall effects (Table \ref{tbl:by-genre}). All genres agree on male associations for \textit{strong}, \textit{active}, \textit{beauty}, and \textit{dominant}, and female associations for \textit{weak}, \textit{submissive}, \textit{childish}, \textit{afraid}, \textit{dependent}, and \textit{hysterical}. For the other stereotypes, we see a handful of genres where the gender association runs contrary to the overall effect. For example, men are more associated with the \textit{emotional} stereotype in Action and Science Fiction, and women are more associated with the \textit{sexual} stereotype in Adventure and Historical Fiction.

Investigating individual words within our stereotype categories, we find 409 significant associations, of which 222 are associated with female characters, and 187 with male characters (Figure \ref{fig:wordcloud}). While the significant words for male characters are more or less evenly divided between verbs (53\%) and adjectives (47\%), the significant words for female characters consist mostly of adjectives (72\%).

\begin{table}[tb]\scriptsize
\renewcommand{\arraystretch}{1.2}
  \begin{tabular}{p{10em}@{\hspace{1em}}p{3em}@{\hspace{4em}}p{10em}@{\hspace{1em}}p{3em}}
  \textbf{Positive with rating} & \textbf{Coef.} & \textbf{Negative with rating} & \textbf{Coef.} \\
  \hline
  sexual (male) & 2.03 & strong (female) & -0.96 \\
  arrogant (male) & 1.45 & domestic (male) & -0.66 \\
  sexual (female) & 1.24 & afraid (female) & -0.66 \\
  violence (male) & 0.92 & weak (male) & -0.63 \\
  active (male) & 0.90 & domestic (female) & -0.57 \\
  hysterical (male) & 0.59 & strong (male) & -0.51 \\
  hysterical (female) & 0.57 & dominant (female) & -0.44 \\
  anger (male) & 0.56 & emotional (female) & -0.44 \\
  violence (female) & 0.46 & beautiful (female) & -0.39 \\
  childish (male) & 0.42 & weak (female) & -0.34 \\
  angry (female) & 0.20 & dependent (female) & -0.25 \\
  emotional (male) & 0.12 & childish (female) & -0.21 \\
  submissive (female) & 0.02 & active (female) & -0.03 \\

\end{tabular}
\caption{Thirty out of thirty-two categories of stereotype are significantly associated ($p < 0.0016$) with positive and negative story ratings in a logistic regression. Sexual and arrogant men are the strongest predictors of a high rated story. Strong women and domestic men are the strongest predictors of a low rated story.}
  \label{tbl:coef}
\end{table}

\begin{figure}[!t]
\centering
\includegraphics[width=1.0\columnwidth]{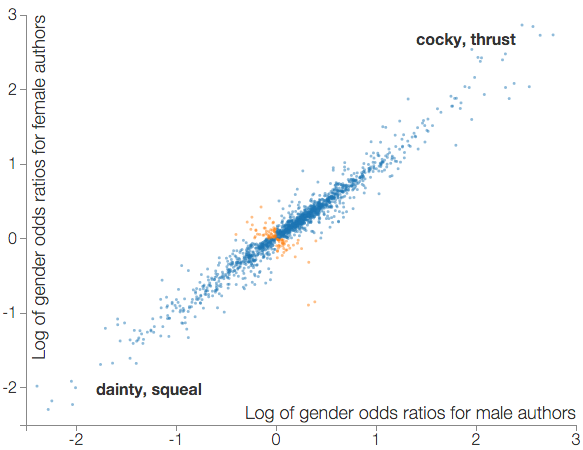}
\caption{Male and female authors write very similar characters, with 0.98 Pearson correlation ($p < 0.001$) over the odds ratios of verbs and adjectives ascribed to male and female characters. In blue we plot the odds ratios that hold the same direction across male and female authors (94\%), and in orange the ones that shift direction (6\%). The lower left quadrant consists of verbs and adjectives both men and women ascribe to female characters (e.g. dainty, squeal). The upper right quadrant consists of verbs and adjectives that both men and women ascribe to male characters (e.g. cocky, thrust).}
\label{fig:scatter}
\end{figure}

\subsection{RQ3: How do stereotypes affect story ratings?}
A logistic regression model that we trained on normalized stereotype counts for each character gender predicts story rating at an accuracy of 88\%. As this model makes predictions over two classes (stories above and below the median rating), chance accuracy is 50\%. In other words, gender stereotypes are strongly associated with story ratings. We present the categories of stereotype that are statistically significant for this model in Figure \ref{tbl:coef}. 

\textbf{Positive associations with rating}. \textit{Sexual}, \textit{violent}, \textit{hysterical}, and \textit{angry} stereotypes are positively associated with story rating for both male and female characters. \textit{Arrogant}, \textit{active}, \textit{childish}, and \textit{emotional} stereotypes are also positively associated, but only for male characters. \textit{Submissiveness} is positively associated with rating, but only for female characters.

\textbf{Negative associations with rating}.
\textit{Domestic}, \textit{strong}, and \textit{weak} stereotypes are negatively associated with story rating for both male and female characters. \textit{Afraid}, \textit{dominant}, \textit{emotional}, \textit{beautiful}, \textit{dependent}, and \textit{active} stereotypes are also negatively associated, but only for female characters.

\subsection{RQ4: Do men and women write genders differently?}
A logistic regression model that we trained on normalized stereotype counts for each gender predicts author gender at an accuracy of 53\% (chance accuracy is 50\%). If male and female authors wrote different kinds of gender stereotypes in their stories, this model would perform much higher than chance. So, it appears that both men and women write indistinguishably stereotypical genders. While this model provides little practical predictive value, one feature of the model did achieve significance after correction: the \textit{active} stereotype for female characters is weakly associated with male authors. This means that men are slightly more likely to write female characters who take physical actions (verbs like ``chase'' or ``lift'') in their stories.

In our second analysis, we find a Pearson correlation of 0.98 over odds ratios that describe the likelihoods of character actions and adjective descriptors written by male and female authors. 94\% of these odds ratios agree in their direction (Figure \ref{fig:scatter}). For example, both male and female authors associate \textit{punching} with male characters, with 2.13 and 2.0 odds respectively. Alternatively, a small number (6\%) of verbs and adjectives shift direction between male and female authors. For example, male authors are more likely to apply the adjective ``immortal'' to female characters, whereas female authors are more likely to apply it to male characters. The data subsets for male and female authors uniformly support the high level relationships in Table \ref{tbl:code-stats}, with slight deviations in the magnitudes of these trends. For example, female authors are slightly less likely to apply hysterical adjectives to female characters than male authors, but overall still quite likely to do so.

\section{Discussion}
We asked whether amateur writers, limited only by their imaginations, might contradict conventional gender stereotypes. To this question, it seems the answer is largely no. We also investigated whether (forward-thinking) genres or author genders are more likely to buck the trend. The answer, again, seems to be no.

\subsection{RQ1-Behavior: How do characters act?}
The statistically strongest male-oriented actions confirm gender stereotypes like \textit{arrogant}, \textit{active}, \textit{violent}, \textit{angry}, and \textit{sexual}. Male characters more often \textit{smirk} (2.7 odds) and \textit{mock} (1.6 odds), \textit{dodge} (1.5 odds) and \textit{chase} (1.6 odds), \textit{abuse} (4.4) and \textit{kill} (1.5), \textit{roar} (4.3 odds) and \textit{curse} (1.6 odds), or \textit{kiss} (2.4 odds) and \textit{lick} (2.0 odds).  For example, ``he narrowed his eyes and \textbf{mocked} her laughter,'' ``he \textbf{dodged} attack after attack,'' ``he mentally \textbf{cursed},'' or ``he tried to \textbf{lick} her ear.'' Similarly, the statistically strongest female-oriented actions reinforce stereotypes through categories like \textit{childish}, \textit{hysterical}, \textit{submissive}, \textit{afraid} or \textit{domestic}. Female characters more often \textit{giggle} (6.7 odds) and \textit{bounce} (2.7 odds), \textit{squeal} (10.3 odds) and \textit{stomp} (2.0 odds), \textit{cook} (2.3 odds) and \textit{wash} (1.8 odds), \textit{cling} (4.0 odds) and \textit{whimper} (2.3 odds), or \textit{shiver} (3.2 odds) and \textit{scream} (2.2 odds). For example, ``she \textbf{bounced} up and down in excitement,'' ``she would help \textbf{clean} the compound with all other women in the clan,'' or ``she \textbf{clung} to him for a while.''

While none of our high-level stereotype categories run directly against behavioral stereotypes, we do observe a surprisingly weak male association with \textit{angry} (1.05 odds). Despite the ``angry man'' stereotype, many common behavioral terms in this category are associated with female characters, for example, \textit{stomp} (1.7 odds), \textit{huff} (1.4 odds), \textit{force} (1.3 odds), and \textit{storm} (1.3 odds). In contrast, the male associated behaviors in this category tend to have a darker connotation, like \textit{beat} (2.0 odds), \textit{rip} (1.7 odds), or \textit{snarl} (1.3 odds).
 
 
 \subsection{RQ2-Description: How are characters described?}
The most statistically male-oriented adjectives confirm gender stereotypes through associations like \textit{strong}, \textit{arrogant}, and \textit{dominant}. Male characters are more often described as \textit{intense} (3.1 odds) and \textit{intimidating} (2.1 odds),  \textit{cocky} (7.1 odds) and \textit{rude} (1.4 odds), or \textit{rich} (2.8 odds) and \textit{famous} (2.5 odds). For example, ``this guy was so \textbf{intimidating},'' ``he was a typical \textbf{cocky} Californian guy,'' or ``his \textbf{rich} and \textbf{famous} dad.'' Likewise, the most statistically female-oriented adjectives  reinforce stereotypes like \textit{submissive}, \textit{dependent}, \textit{hysterical}.  Female characters are more often described as  \textit{helpless} (2.4 odds) and \textit{fragile} (5.9 odds), \textit{clingy} (4.4 odds) and \textit{vulnerable} (2.1 odds), or \textit{bitchy} (12.2 odds) and \textit{dramatic} (2.1 odds). For example, ``she went \textbf{weak} in the knees,'' or ``like all overly \textbf{dramatic} teenage girls, Vanny got over her breakup,'' or ``she was too \textbf{clingy}, so he dumped her.''

Against traditional gender stereotypes, we find that \textit{beautiful} is significantly male associated. This effect size is weak, however, and both genders are often physically objectified in our corpus. Women are more often described as \textit{beautiful} (3.3 odds), \textit{flawless} (1.6 odds), \textit{curvy} (46.0 odds), or \textit{slim} (3.5 odds). For example, ``her pale skin was nearly \textbf{flawless}.'' But men are described with their own comparable set of terms, including \textit{gorgeous} (1.3 odds), \textit{hot} (2.2 odds), \textit{dreamy} (3.0 odds), and \textit{toned} (19.3 odds). For example, ``she watched him walk, \textbf{shirtless} and dangerously \textbf{gorgeous}.'' 

\subsection{RQ1 and RQ2: Stereotypes among genres}
Gender stereotypes are mostly consistent across genres, suggesting they are not driven by any one particularly popular sub-community on Wattpad. For example, when we first discovered the male association with \textit{beauty}, we wondered whether slash fiction (a popular sub-genre of Fanfiction focused on gay relationships) might have biased our results. The genre analysis provides strong evidence to the contrary, as the \textit{beauty} association held across every genre in our corpus. Similarly, we find that ten of the fifteen stereotype categories we analyze are completely consistent across genres.

Beyond providing evidence that stereotypes generalize across Wattpad, the genre analysis opens up a series of new questions. For example, why are women more associated with \textit{sexual} stereotypes in General Fiction, Science Fiction, Historical Fiction, and Adventure stories? Does this deviation give female characters more or less agency than we see in other genres? 
Similarly, why are men more \textit{emotional} among Action, Adventure, and Science Fiction stories? On inspection, it appears that strong emotions emerge in the course and aftermath of physical conflict, disproportionately undertaken by male characters in these genres. Better characterizing these sub-communities remains future work. 

\subsection{RQ3: Story rating and stereotypes}
While gendered category counts strongly predict story rating, conventional stereotypes are not always associated with higher rated stories. In support of conventional stereotypes, \textit{sexual}, \textit{arrogant}, \textit{violent}, \textit{angry}, and \textit{active} men and \textit{hysterical}, \textit{submissive} women are positively associated with story rating.  Similarly, \textit{domestic}, \textit{weak} men and \textit{strong}, \textit{dominant} women are negatively associated with rating. However, these results are tempered by contrasting trends. \textit{Sexual}, \textit{violent}, and \textit{angry} women are also associated with higher rated stories (albeit with weaker coefficients than their male counterparts), as are  \textit{childish} and \textit{emotional} men.

Four stereotype categories are significant for only one gender. \textit{Afraid}, \textit{dependent}, and \textit{beautiful} female characters are associated with lower rated stories, and \textit{arrogant} male characters are associated with higher rated stories. Similarly, two stereotype categories flip between genders. \textit{Active} and \textit{childish} male characters are associated with higher rated stories, whereas \textit{active} and \textit{childish} female characters are associated with lower rated stories.

Taken as a whole, these results present a mixed account of the how  gender stereotypes are received on Wattpad. Clearly, they exist in the community. But they may not always be encouraged by readers. Many common stereotypes, like \textit{sexual} and \textit{violent} men, have a positive impact on rating, but others, like \textit{beautiful} women, have a negative impact.

\subsection{RQ4-Authors: Author gender and stereotypes}
Authors of both genders are responsible for the stereotypes we see in the Wattpad dataset. Our first analysis suggests that patterns of gender stereotypes in a story cannot be used to predict author gender.
And our second analysis shows that men and women write extremely similar kinds of characters. Of the 2000 verbs and adjectives we analyzed for association with male and female characters, few of them disagree between author genders (the orange dots in Figure \ref{fig:scatter}). 

Of these points of disagreement between male and female authors, 40\% involve words in our stereotype categories. A handful of these words present instances in which women reject gender stereotypes. For example, female authors are more likely to apply the adjective ``strongest'' to female characters, and ``weary'' to male characters, while male authors do the opposite. Others of these words speak to a sense of difference between genders that lies outside conventional stereotypes. One group of the statistics is self-deprecating within gender. For example, men are more likely to describe male characters as ``flustered,'' ``embarrassed,'' or ``chubby,'' while women are more likely to apply the same adjectives to female characters.  Another group of the statistics aggrandizes the opposite gender. For example, women are more likely to describe male characters as ``smarter,'' ``magical,'' or ``sparkling,'' while men more often apply these adjectives to female characters.

\section{Limitations}
Our approach has limitations, which in turn provide opportunities for future work.

\textbf{Sentence level analysis}. In this paper, we analyze language at the level of sentences, examining verbs and adjectives for bias based on gender. While the concreteness of this approach allows us to efficiently draw out quantitative insight, it also excludes the greater context of the story. An independent female character \textit{dying} after leading a rebellion, for example, tells us something different about gender bias than the death of a submissive and beautiful female character.  
Looking beyond the sentence level to story structure could allow us to consider how stereotypes interact with each other, or relate to earlier plot events in a story.

\textbf{External Validity.} Our work asks whether modern writers, free from market incentives, would reinforce or contradict gender stereotypes in their stories. There are several ways in which our results may not generalize to fully answer this question. First, while the Wattpad writing community is worth studying in its own right, it may not accurately represent the broader community of writers, amateur or otherwise. Second, writers on Wattpad may not be entirely free of commercial incentives: some writers have used the community as a launching point for their careers in commercial fiction, and many writers desire as wide an audience as possible. However, given the immense scale and diversity of the Wattpad dataset (more than 1.8 billion words, split almost evenly across male and female authors), we suggest that it provides the best available means towards answering our questions.

\section{Conclusion}
We ask whether modern amateur writers, limited only by their imaginations, reinforce or contradict gender stereotypes. To find out, we analyze more than 1.8 billion words of  fiction on Wattpad, an online writing community. We find that modern amateur writers do reinforce traditional stereotypes, and that male and female authors are just as likely to perpetuate them.

\section{Acknowledgments}
First, we would like to thank Jordan Christensen and Mohammad Islam at Wattpad -- this work wouldn't be possible without their help and collaboration. We would also like to thank our reviewers and colleagues at Stanford for their helpful feedback. This work is supported by a NSF Graduate Fellowship.

\bibliographystyle{aaai}
\bibliography{ref}

\begin{thebibliography}{}

\bibitem[\protect\citeauthoryear{Bergstrom, Jenson, and de
  Castell}{2012}]{roleplayinggames}
Bergstrom, K.; Jenson, J.; and de~Castell, S.
\newblock 2012.
\newblock What's 'choice' got to do with it?: Avatar selection differences
  between novice and expert players of world of warcraft and rift.

\bibitem[\protect\citeauthoryear{Bertrand and Hallock}{2001}]{industry}
Bertrand, M., and Hallock, K.~F.
\newblock 2001.
\newblock The gender gap in top corporate jobs.
\newblock {\em Industrial \& Labor Relations Review} 55(1):3--21.

\bibitem[\protect\citeauthoryear{Bretthauer, Zimmerman, and
  Banning}{2007}]{music}
Bretthauer, B.; Zimmerman, T.~S.; and Banning, J.~H.
\newblock 2007.
\newblock A feminist analysis of popular music: Power over, objectification of,
  and violence against women.
\newblock {\em Journal of Feminist Family Therapy} 18(4):29--51.

\bibitem[\protect\citeauthoryear{Emons, Wester, and Scheepers}{2010}]{dutchtv}
Emons, P.; Wester, F.; and Scheepers, P.
\newblock 2010.
\newblock He works outside the home; she drinks coffee and does the dishes:
  Gender roles in fiction programs on dutch television.

\bibitem[\protect\citeauthoryear{Fast, Chen, and Bernstein}{2016}]{empath}
Fast, E.; Chen, B.; and Bernstein, M.
\newblock 2016.
\newblock Empath: Understanding topic signals in large-scale text.
\newblock {\em In Proc. CHI 2016}.

\bibitem[\protect\citeauthoryear{Fiore \bgroup et al\mbox.\egroup
  }{2008}]{datingprofiles}
Fiore, A.~T.; Taylor, L.~S.; Mendelsohn, G.; and Hearst, M.
\newblock 2008.
\newblock Assessing attractiveness in online dating profiles.
\newblock {\em In Proc. CHI 2008}.

\bibitem[\protect\citeauthoryear{Garcia, Weber, and Garimella}{2014}]{bechdel}
Garcia, D.; Weber, I.; and Garimella, V.
\newblock 2014.
\newblock Gender asymmetries in reality and fiction: The bechdel test of social
  media.
\newblock {\em ICWSM 2014}.

\bibitem[\protect\citeauthoryear{Gilbert \bgroup et al\mbox.\egroup
  }{2013}]{pinterest}
Gilbert, E.; Bakhshi, S.; Chang, S.; and Terveen, L.
\newblock 2013.
\newblock "i need to try this"?: A statistical overview of pinterest.
\newblock {\em In Proc. CHI 2013}.

\bibitem[\protect\citeauthoryear{Gilbert}{2012}]{enron}
Gilbert, E.
\newblock 2012.
\newblock Phrases that signal workplace hierarchy.
\newblock {\em In Proc. CSCW 2012}.

\bibitem[\protect\citeauthoryear{Gooden and Gooden}{2001}]{children}
Gooden, A.~M., and Gooden, M.~A.
\newblock 2001.
\newblock Gender representation in notable children's picture books:
  1995--1999.
\newblock {\em Sex Roles} 45(1-2):89--101.

\bibitem[\protect\citeauthoryear{Jia, Lansdall-Welfare, and
  Cristianini}{2015}]{newsimages}
Jia, S.; Lansdall-Welfare, T.; and Cristianini, N.
\newblock 2015.
\newblock Measuring gender bias in news images.
\newblock {\em In Proc. WWW 2015}.

\bibitem[\protect\citeauthoryear{Kay, Matuszek, and Munson}{2015}]{imagesearch}
Kay, M.; Matuszek, C.; and Munson, S.~A.
\newblock 2015.
\newblock Unequal representation and gender stereotypes in image search results
  for occupations.
\newblock {\em In Proc. CHI 2015}.

\bibitem[\protect\citeauthoryear{Kramer, Guillory, and
  Hancock}{2014}]{contagion}
Kramer, A. D.~I.; Guillory, J.~E.; and Hancock, J.~T.
\newblock 2014.
\newblock Experimental evidence of massive-scale emotional contagion through
  social networks.
\newblock {\em Proceedings of the National Academy of Sciences}
  111(24):8788--8790.

\bibitem[\protect\citeauthoryear{Lauzen, Dozier, and Horan}{2008}]{primetime}
Lauzen, M.; Dozier, D.; and Horan, N.
\newblock 2008.
\newblock Constructing gender stereotypes through social roles in prime-time
  television.

\bibitem[\protect\citeauthoryear{Mohammad and Turney}{2013}]{emolex}
Mohammad, S.~M., and Turney, P.~D.
\newblock 2013.
\newblock Crowdsourcing a word-emotion association lexicon.
\newblock {\em Computational Intelligence} 29(3):436--465.

\bibitem[\protect\citeauthoryear{Otterbacher}{2013}]{imdb}
Otterbacher, J.
\newblock 2013.
\newblock Gender, writing and ranking in review forums: a case study of the
  imdb.
\newblock {\em Knowledge and information systems} 35(3):645--664.

\bibitem[\protect\citeauthoryear{Otterbacher}{2015}]{descriptionbias}
Otterbacher, J.
\newblock 2015.
\newblock Crowdsourcing stereotypes: Linguistic bias in metadata generated via
  gwap.
\newblock {\em In Proc. CHI 2015}.

\bibitem[\protect\citeauthoryear{Ottoni \bgroup et al\mbox.\egroup
  }{2013}]{pinterest2}
Ottoni, R.; Pesce, J.~P.; Las~Casas, D.~B.; Franciscani~Jr, G.; Meira~Jr, W.;
  Kumaraguru, P.; and Almeida, V.
\newblock 2013.
\newblock Ladies first: Analyzing gender roles and behaviors in pinterest.
\newblock {\em In Proc. ICWSM}.

\bibitem[\protect\citeauthoryear{Pennebaker and Stone}{2003}]{wisdom}
Pennebaker, J.~W., and Stone, L.~D.
\newblock 2003.
\newblock Words of wisdom: language use over the life span.
\newblock {\em Journal of personality and social psychology} 85(2):291.

\bibitem[\protect\citeauthoryear{Pennebaker, Francis, and Booth}{2001}]{liwc}
Pennebaker, J.~W.; Francis, M.~E.; and Booth, R.~J.
\newblock 2001.
\newblock Linguistic inquiry and word count: Liwc 2001.
\newblock {\em Mahway: Lawrence Erlbaum Associates 71 2001}.

\bibitem[\protect\citeauthoryear{Rose \bgroup et al\mbox.\egroup
  }{2012}]{facebookpictures}
Rose, J.; Mackey-Kallis, S.; Shyles, L.; Barry, K.; Biagini, D.; Hart, C.; and
  Jack, L.
\newblock 2012.
\newblock Face it: The impact of gender on social media images.
\newblock {\em Communication Quarterly 2012}.

\bibitem[\protect\citeauthoryear{Ross and Carter}{2011}]{news}
Ross, K., and Carter, C.
\newblock 2011.
\newblock Women and news: A long and winding road.
\newblock {\em Media, Culture \& Society} 33(8):1148--1165.

\bibitem[\protect\citeauthoryear{Salehi, Irani, and Bernstein}{2015}]{dynamo}
Salehi, N.; Irani, L.~C.; and Bernstein, M.~S.
\newblock 2015.
\newblock We are dynamo: Overcoming stalling and friction in collective action
  for crowd workers.
\newblock {\em Proceedings of the 33rd Annual ACM Conference on Human Factors
  in Computing Systems}  1621--1630.

\bibitem[\protect\citeauthoryear{Sheng, Provost, and
  Ipeirotis}{2008}]{get-another-label}
Sheng, V.~S.; Provost, F.; and Ipeirotis, P.~G.
\newblock 2008.
\newblock Get another label? improving data quality and data mining using
  multiple, noisy labelers.
\newblock {\em Proceedings of the 14th ACM SIGKDD international conference on
  Knowledge discovery and data mining}  614--622.

\bibitem[\protect\citeauthoryear{Soulliere}{2006}]{wrestling}
Soulliere, D.
\newblock 2006.
\newblock Wrestling with masculinity: Messages about manhood in the wwe.

\bibitem[\protect\citeauthoryear{Sugimoto}{2013}]{science}
Sugimoto, C.~R.
\newblock 2013.
\newblock Global gender disparities in science.
\newblock {\em Nature} 504(7479):211--213.

\bibitem[\protect\citeauthoryear{Towbin \bgroup et al\mbox.\egroup
  }{2008}]{disneyfilms}
Towbin, M.~A.; Haddock, S.~A.; Zimmerman, T.~S.; Lund, L.~K.; and Tanner, L.~R.
\newblock 2008.
\newblock Images of gender, race, age, and sexual orientation in disney
  feature-length animatedfilms.
\newblock {\em Journal of Feminist Family Therapy 2008}.

\bibitem[\protect\citeauthoryear{Wagner \bgroup et al\mbox.\egroup
  }{2015}]{wikipedia}
Wagner, C.; Garcia, D.; Jadidi, M.; and Strohmaier, M.
\newblock 2015.
\newblock It's a man's wikipedia? assessing gender inequality in an online
  encyclopedia.
\newblock {\em In Proc. ICWSM 2015}.

\end{thebibliography}

\end{document}